\documentclass[letterpaper]{article}

\usepackage{times}
\usepackage{helvet}
\usepackage{courier}
\usepackage{url}
\usepackage{graphicx}
\usepackage[utf8]{inputenc} 
\usepackage[T1]{fontenc}    
\usepackage{booktabs}       
\usepackage{amsfonts}       
\usepackage{nicefrac}       
\usepackage{microtype}      
\usepackage{tikz}
\usetikzlibrary{shapes.misc, positioning}
\usepackage{thmtools,thm-restate}
\usepackage{parskip}
\usepackage[noblocks]{authblk}

\usepackage{floatrow}
\newfloatcommand{capbtabbox}{table}[][\FBwidth]

\usepackage{basic}

\title{MLSys: The New Frontier of Machine Learning Systems}

\makeatletter
\renewcommand\AB@affilsepx{, \protect\Affilfont}
\makeatother

\renewcommand\Affilfont{\small}

\author[1,2,3]{Alexander Ratner}
\author[4]{Dan Alistarh}
\author[5]{Gustavo Alonso}
\author[6,7]{David G. Andersen}
\author[1,8]{Peter Bailis}
\author[9]{Sarah Bird}
\author[7]{Nicholas Carlini}
\author[10]{Bryan Catanzaro}
\author[9]{Jennifer Chayes}
\author[9]{Eric Chung}
\author[1,10]{Bill Dally}
\author[7]{Jeff Dean}
\author[11,12]{Inderjit S. Dhillon}
\author[11]{Alexandros Dimakis}
\author[13]{Pradeep Dubey}
\author[14]{Charles Elkan}
\author[15,16]{Grigori Fursin}
\author[6]{Gregory R. Ganger}
\author[17]{Lise Getoor}
\author[6]{Phillip B. Gibbons}
\author[18,19,6]{Garth A. Gibson}
\author[20]{Joseph E. Gonzalez}
\author[13]{Justin Gottschlich}
\author[21]{Song Han}
\author[22]{Kim Hazelwood}
\author[23]{Furong Huang}
\author[24]{Martin Jaggi}
\author[2]{Kevin Jamieson}
\author[20]{Michael I. Jordan}
\author[6]{Gauri Joshi}
\author[25]{Rania Khalaf}
\author[13]{Jason Knight}
\author[7]{Jakub Konečný}
\author[21]{Tim Kraska}
\author[14]{Arun Kumar}
\author[26]{Anastasios Kyrillidis}
\author[22]{Aparna Lakshmiratan}
\author[27]{Jing Li }
\author[21]{Samuel Madden}
\author[7]{H. Brendan McMahan}
\author[22]{Erik Meijer}
\author[28,29]{Ioannis Mitliagkas}
\author[7]{Rajat Monga}
\author[7]{Derek Murray}
\author[1,30]{Kunle Olukotun}
\author[27]{Dimitris Papailiopoulos}
\author[31]{Gennady Pekhimenko}
\author[1]{Christopher R\'e}
\author[27]{Theodoros Rekatsinas}
\author[7]{Afshin Rostamizadeh}
\author[32]{Christopher De Sa}
\author[7]{Hanie Sedghi}
\author[9]{Siddhartha Sen}
\author[6]{Virginia Smith}
\author[12,6]{Alex Smola}
\author[20]{Dawn Song}
\author[33]{Evan Sparks}
\author[20]{Ion Stoica}
\author[21]{Vivienne Sze}
\author[32]{Madeleine Udell}
\author[34]{Joaquin Vanschoren}
\author[27]{Shivaram Venkataraman}
\author[6]{Rashmi Vinayak}
\author[9]{Markus Weimer}
\author[32]{Andrew Gordon Wilson}
\author[6,35]{Eric Xing}
\author[1,36]{Matei Zaharia}
\author[5]{Ce Zhang}
\author[6,33]{Ameet Talwalkar\footnote{Corresponding author, talwalkar@cmu.edu.}}

\affil[1]{Stanford}
\affil[2]{University of Washington}
\affil[3]{Snorkel AI}
\affil[4]{IST Austria}
\affil[5]{ETH Zurich}
\affil[6]{Carnegie Mellon University}
\affil[7]{Google}
\affil[8]{Sisu Data}
\affil[9]{Microsoft}
\affil[10]{NVIDIA}
\affil[11]{University of Texas at Austin}
\affil[12]{Amazon}
\affil[13]{Intel}
\affil[14]{University of California San Diego}
\affil[15]{cTuning Foundation}
\affil[16]{Dividiti}
\affil[17]{UC Santa Cruz}
\affil[18]{Vector Institute}
\affil[19]{Univerrsity of Toronto}
\affil[20]{UC Berkeley}
\affil[21]{MIT}
\affil[22]{Facebook}
\affil[23]{University of Maryland}
\affil[24]{EPFL}
\affil[25]{IBM Research}
\affil[26]{Rice University}
\affil[27]{University of Wisconsin-Madison}
\affil[28]{Mila}
\affil[29]{University of Montreal}
\affil[30]{SambaNova Systems}
\affil[31]{University of Toronto}
\affil[32]{Cornell University}
\affil[33]{Determined AI}
\affil[34]{Eindhoven University of Technology}
\affil[35]{Petuum}
\affil[36]{Databricks}

\begin{document}

\date{May 2, 2019}

\maketitle

\begin{abstract}
Machine learning (ML) techniques are enjoying rapidly increasing adoption.
However, designing and implementing the systems that support ML models in real-world deployments remains a significant obstacle, in large part due to the radically different development and deployment profile of modern ML methods, and the range of practical concerns that come with broader adoption.
We propose to foster a new systems machine learning research community at the intersection of the traditional systems and ML communities, focused on topics such as hardware systems for ML, software systems for ML, and ML optimized for metrics beyond predictive accuracy.
To do this, we describe a new conference, MLSys, that explicitly targets research at the intersection of systems and machine learning with a program committee split evenly between experts in systems and ML, and an explicit focus on topics at the intersection of the two.
\end{abstract}

\section{Introduction}
\label{sec:intro}
Over the last few years, machine learning (ML) has hit an inflection point in terms of adoption and results.
Large corporations have invested billions of dollars in reinventing themselves as “AI-centric”; swaths of academic disciplines have flocked to incorporate machine learning into their research; and a wave of excitement about AI and ML has proliferated through the broader public sphere.
This has been due to several factors, central amongst them new deep learning approaches, increasing amounts of data and compute resources, and collective investment in open-source frameworks like Caffe, Theano, MXNet, TensorFlow, and PyTorch, which have effectively decoupled model design and specification from the systems to implement these models.
The resulting wave of technical advances and practical results seems poised to transform ML from a bespoke solution used on certain narrowly-defined tasks, to a commodity technology deployed nearly everywhere.

Unfortunately, while it is easier than ever to run state-of-the-art ML models on pre-packaged datasets, designing and implementing the \textit{systems} that support ML in real-world applications is increasingly a major bottleneck.
In large part this is because ML-based applications require distinctly new types of software, hardware, and engineering systems to support them.
Indeed, modern ML applications have been referred to by some as a new “Software 2.0”~\cite{karpathy2018software} to emphasize the radical shift they represent as compared to traditional computing applications.
They are increasingly \textit{developed} in different ways than traditional software---for example, by collecting, preprocessing, labeling, and reshaping training datasets rather than writing code---and also \textit{deployed} in different ways, for example utilizing specialized hardware, new types of quality assurance methods, and new end-to-end workflows.
This shift opens up exciting research challenges and opportunities around high-level interfaces for ML development, low-level systems for executing ML models, and interfaces for embedding learned components in the middle of traditional computer systems code.

Modern ML approaches also require new solutions for the set of concerns that naturally arise as these techniques gain broader usage in diverse real-world settings.
These include \textit{cost} and other efficiency metrics for small and large organizations alike, including e.g. computational cost at training and prediction time, engineering cost, and cost of errors in real-world settings; \textit{accessibility} and \textit{automation}, for the expanding set of ML users that do not have PhDs in machine learning, or PhD time scales to invest; \textit{latency} and other run-time constraints, for a widening range of computational deployment environments; and concerns like \textit{fairness, bias, robustness, security, privacy, interpretability, and causality}, which arise as ML starts to be applied to critical settings where impactful human interactions are involved, like driving, medicine, finance, and law enforcement.

This combination of radically different application requirements, increasingly-prevalent systems-level concerns, and a rising tide of interest and adoption, collectively point to the need for a concerted research focus on the systems aspects of machine learning.
To accelerate these research efforts, our goal is to help foster a new \textit{systems machine learning community} dedicated to these issues.
We envision focusing on broad, full-stack questions that are complementary to those traditionally tackled independently by the ML and Systems communities, including:

\begin{enumerate}
    \item \textbf{How should \highlight{software systems} be designed to support the full machine learning lifecycle, from programming interfaces and data preprocessing to output interpretation, debugging and monitoring? Example questions include:}
    \begin{itemize}
        \item \textit{How can we enable users to quickly “program” the modern machine learning stack through emerging interfaces such as manipulating or labeling training data, imposing simple priors or constraints, or defining loss functions?}
        \item \textit{How can we enable developers to define and measure ML models, architectures, and systems in higher-level ways?}
        \item \textit{How can we support efficient development, monitoring, interpretation, debugging, adaptation, tuning, and overall maintenance of production ML applications- including not just models, but the data, features, labels, and other inputs that define them?}
    \end{itemize}

    \item \textbf{How should \highlight{hardware systems} be designed for machine learning? Example questions include:}
    \begin{itemize}
        \item \textit{How can we develop specialized, heterogeneous hardware for training and deploying machine learning models, fit to their new operation sets and data access patterns?}
        \item \textit{How can we take advantage of the stochastic nature of ML workloads to discover new trade-offs with respect to precision, stability, fidelity, and more?}
        \item \textit{How should distributed systems be designed to support ML training and serving?}
    \end{itemize}

    \item \textbf{How should machine learning systems be designed to satisfy \highlight{metrics beyond predictive accuracy}, such as power and memory efficiency, accessibility, cost, latency, privacy, security, fairness, and interpretability? Example questions include:}
    \begin{itemize}
        \item \textit{How can machine learning algorithms and systems be designed for device constraints such as power, latency, and memory limits?}
        \item \textit{How can ML systems be designed to support full-stack privacy and security guarantees, including, e.g., federated learning and other similar settings?}
        \item \textit{How can we increase the accessibility of ML, to empower an increasingly broad range of users who may be neither ML nor systems experts?}
    \end{itemize}
\end{enumerate}

Another way of partitioning these research topics is into \highlight{high-level systems} for ML that support interfaces and workflows for ML development---the analogue of traditional work on programming languages and software engineering---and \highlight{low-level systems} for ML that involve hardware or software---and that often blur the lines between the two---to support training and execution of models, the analogue of traditional work on compilers and architecture.
Regardless of the ontology, we envision these questions being addressed by a strong mix of theoretical, empirical, and applications-driven perspectives.
And given their full-stack nature, we see them being best answered by a research community that mixes perspectives from the traditional machine learning and systems communities.

A separate but closely related and increasingly exciting area of focus is \textit{machine learning for systems}: the idea of applying machine learning techniques to improve traditional computing systems.
Examples include replacing the data structures, heuristics, or hand-tuned parameters used in low-level systems like operating systems, compilers, and storage systems with learned models.
While this is clearly a distinct research direction, we also see the systems machine learning community as an ideal one to examine and support this line of work, given the required confluence of ML and systems expertise.

Finally, we see the systems machine learning community as an ideal jumping-off point for even larger-scale and broader questions, beyond how to interface with, train, execute, or evaluate single models~\cite{jordan2018}.
For instance, how do we manage entire ecosystems of models that interact in complex ways?
How do we maintain and evaluate systems that pursue long term goals?
How do we measure the effect of ML systems on societies, markets, and more?
How do we share and reuse data and models at societal scale, while maintaining privacy and other economic, social, and legal issues?
All of these questions and many more will likely need to be approached by research at the intersection of traditional machine learning and systems viewpoints.

\section{Why Now? The Rise of Full Stack Bottlenecks in ML}
\label{sec:why-now}
Researchers have worked at the intersection of systems and machine learning research for years---but this \textit{systems ML} work has moved to the forefront recently due to leaps in machine learning performance on challenging benchmark tasks, and the growing realization that a range of new systems up and down the computing stack are needed to translate this academic promise to real-world practice~\cite{talwalkar2018,sculley2014machine}.

In recent years, often driven by new deep learning approaches, the field of machine learning has made significant leaps forward on benchmark tasks in traditional ‘grand challenge’ domains like image classification, text and speech processing, and others.
In certain benchmarks, ML models have even surpassed human performance~\cite{eckersley2017eff}.
However, in real-world deployments, a range of bottlenecks begin to surface, which crucially are full-stack, systems-level concerns, rather than solely properties of the core machine learning algorithms.
These include:

\begin{itemize}
    \item \textbf{Deployment concerns:} As ML becomes used in increasingly diverse and mission-critical ways, a new crop of systems-wide concerns has become increasingly prevalent.
    These include robustness to adversarial influences or other spurious factors; safety more broadly considered; privacy and security, especially as sensitive data is increasingly used; interpretability, as is increasingly both legally and operationally required; fairness, as ML algorithms begin to have major effects on our everyday lives; and many other similar concerns.

    \item \textbf{Cost:}
    The original default solution for learning a CNN over ImageNet cost \$2,300 of compute and took 13 days to train~\cite{coleman2017dawnbench}.
    Annotation of the huge volumes of training data can alone cost hundreds of thousands to millions of dollars.
    Reducing cost measured in terms of other metrics---such as latency or power---is also critical for an expanding range of device and production deployment profiles.

    \item \textbf{Accessibility:} As a widening range of people rush to use ML for actual production purposes---including a new breed of polyglot data scientists trained by large new programs at universities---ML systems need to be usable by developers and organizations without PhD-level machine learning and systems expertise.
\end{itemize}

The shared element in these emerging pain points is that they are full-stack issues that require reasoning not just over core ML algorithms and methods, but the hardware, software, and overall systems that support them as well.
As ML adoption and excitement increases, full-stack solutions to the above pain points will be increasingly critical to bridging the gap between machine learning’s promise and real-world utility.

\section{MLSys: Building a New Conference at the Intersection of Systems + Machine Learning}
\label{sec:mlsys}
In our view, these fundamental gaps between ML’s current promise and its actual usability in practice all concern questions at the intersection of the traditional systems and machine learning communities.
As such, we have created a new conference, called the Conference on Machine Learning and Systems (MLSys), to target research at the intersection of systems and machine learning, and with the hope of providing an intellectual space that fosters interdisciplinary research that cuts across both.
The conference aims to elicit new connections amongst these fields, including identifying best practices and design principles for machine learning systems, as well as developing novel learning methods and theory tailored to practical machine learning workflows. 

MLSys builds on the dramatic success and growth of satellite workshops connected to leading ML and Systems conferences, e.g., at NeurIPS, ICML, OSDI, and SIGMOD.
Following the standards of top conferences in both fields, the MLSys review process is a rigorous, highly selective process.
However, unlike traditional ML or Systems conferences, the MLSys Program Committee consists of experts from both ML and Systems who review all papers together, and has an explicit focus on topics that fall in the interdisciplinary MLSys space (as opposed to the broad ML or broad Systems space).
Finally, to spur reproducibility and rapid progress in this research area, MLSys embraces modern artifact evaluation processes that have been successful at other conferences~\cite{acm2018artifact}.

MLSys was established in 2018 by the inaugural Organizing Committee (Peter Bailis, Sarah Bird, Dimitris Papailiopoulos, Chris Ré, Ben Recht, Virginia Smith, Ameet Talwalkar, Matei Zaharia), led by Program Chair Ameet Talwalkar and Co-Chair Dimitris Papailiopoulos, and with guidance from the Steering and Program Committees.
\begin{itemize}
    \item \textit{Steering Committee:} Jennifer Chayes, Bill Dally, Jeff Dean, Michael I. Jordan, Yann LeCun, Fei-Fei Li, Alex Smola, Dawn Song, Eric Xing.
    \item \textit{Program Committee:} David Andersen, Bryan Catanzaro, Eric Chung, Christopher De Sa, Inderjit Dhillon, Alex Dimakis, Charles Elkan, Greg Ganger, Lise Getoor, Phillip Gibbons, Garth Gibson, Joseph Gonzalez, Furong Huang, Kevin Jamieson, Yangqing Jia, Rania Khalaf, Jason Knight, Tim Kraska, Aparna Lakshmiratan, Samuel Madden, Brendan McMahan, Ioannis Mitliagkas, Rajat Monga, Derek Murray, Kunle Olukotun, Theodoros Rekatsinas, Afshin Rostamizadeh, Siddhartha Sen, Evan Sparks, Ion Stoica, Shivaram Venkataraman, Rashmi Vinayak, Markus Weimer, Ce Zhang.
\end{itemize}

\section{Conclusion}
\label{sec:conclusion}
There is an incredibly exciting set of research challenges that can be uniquely tackled at the intersection of traditional machine learning and systems communities, both today and moving forward.
Solving these challenges will require advances in theory, algorithms, software, and hardware, and will lead to exciting new low-level systems for executing ML algorithms, high-level systems for specifying, monitoring, and interacting with them, and beyond that, new paradigms and frameworks that shape how machine learning interacts with society in general.
We envision the new MLSys conference as a center of research in these increasingly important areas.

\bibliography{mlsys-whitepaper}

\begin{thebibliography}{1}

\bibitem{acm2018artifact}
{ACM}: Artifact reviewing and badging.
\newblock
  \url{https://www.acm.org/publications/policies/artifact-review-badging},
  2018.

\bibitem{coleman2017dawnbench}
C.~Coleman, D.~Narayanan, D.~Kang, T.~Zhao, J.~Zhang, L.~Nardi, P.~Bailis,
  K.~Olukotun, C.~R{\'e}, and M.~Zaharia.
\newblock {DAWNBench}: An end-to-end deep learning benchmark and competition.
\newblock {\em NeurIPS ML Systems Workshop}, 2017.

\bibitem{eckersley2017eff}
P.~Eckersley, Y.~Nasser, et~al.
\newblock {EFF AI} progress measurement project.
\newblock \url{https://eff.org/ai/metrics}, 2017.

\bibitem{jordan2018}
M.~Jordan.
\newblock Artificial intelligence---the revolution hasn't happened yet.
\newblock 2018.
\newblock
  \url{https://medium.com/@mijordan3/artificial-intelligence-the-revolution-hasnt-happened-yet-5e1d5812e1e7}.

\bibitem{karpathy2018software}
A.~Karpathy.
\newblock Software 2.0.
\newblock \url{https://medium.com/@karpathy/software-2-0-a64152b37c35}, 2017.

\bibitem{sculley2014machine}
D.~Sculley, G.~Holt, D.~Golovin, E.~Davydov, T.~Phillips, D.~Ebner,
  V.~Chaudhary, and M.~Young.
\newblock Machine learning: The high interest credit card of technical debt.
\newblock 2014.

\bibitem{talwalkar2018}
A.~Talwalkar.
\newblock Toward the jet age of machine learning.
\newblock {\em O'Reilly}, 2018.
\newblock
  \url{https://www.oreilly.com/ideas/toward-the-jet-age-of-machine-learning}.

\end{thebibliography}
\bibliographystyle{abbrv}
\end{document}